\documentclass{article} 
\usepackage{iclr2024_conference,times}
\usepackage[pdftex]{graphicx}
\usepackage{subcaption}

\usepackage{amsmath,amsfonts,bm}

\def\eqref#1{equation~\ref{#1}}

\def\1{\bm{1}}

\DeclareMathAlphabet{\mathsfit}{\encodingdefault}{\sfdefault}{m}{sl}
\SetMathAlphabet{\mathsfit}{bold}{\encodingdefault}{\sfdefault}{bx}{n}

\usepackage{hyperref}
\usepackage{url}

\title{Task agnostic architecture for algorithm induction via implicit composition}

\author{Sahil J. Sindhi 
\thanks{The majority of this work was developed while at the Department of Engineering, University of Cambridge.} \\
Independent researcher \\
\texttt{sahiljsindhi@gmail.com} \\
\And
Ignas Budvytis \footnotemark[1]  \\
Independent researcher \\
\texttt{ignas.budvytis@gmail.com} \\
}

\iclrfinalcopy 
\begin{document}

\maketitle
\begin{abstract}

For many years, different fields in applied machine learning such as computer vision, speech or natural language processing have been building domain-specialised solutions, mainly fueled by the need for hand-crafted features to then be used with classical machine learning methods. Currently, we are witnessing an opposing trend towards developing more generalist architectures, driven by Large Language Models and multi-modal foundational models. These architectures are designed to tackle a variety of tasks, including those that were previously unseen and using inputs across multiple modalities.
Taking this trend of generalization to the extreme suggests the possibility of a single deep network architecture capable of solving all tasks, rather than just one or a limited subset as currently observed. 

This position paper aims to explore the idea that developing such a unified architecture may not be as difficult as one may think and propose a theoretical framework of how it could be constructed. Our proposal is based on the following assumptions. Firstly, the solution to any given task can be expressed through a sequence of instructions, as most tasks necessitate implementation in code on conventional computing hardware, which operates inherently in a sequential manner. Second, recent Generative AI, especially Transformer-based models, demonstrate not only a leap in performance but also a potential as an architecture capable of constructing algorithms for a wide range of domains. For example, GPT-4 shows exceptional capability at in-context learning of novel tasks which is hard to explain in any other way than the ability to compose novel solutions from fragments on previously learnt algorithms.

Third, the observation that the main missing component in developing a truly generalised network is an efficient approach for self-consistent input of previously learnt sub-steps of an algorithm and their (implicit) composition during the network's internal forward pass.

Our exploration delves into the current capabilities and limitations of Transformer-based models and other methods in efficient and correct algorithm composition and proposes a Transformer-like architecture as well as a discrete learning framework to overcome these limitations. 
\end{abstract}

\section{Introduction}

Recently there has been a leap in the quality of machine learning models with the introduction of the Transformer architecture~\citep{transformer}. Most notably Large Language Models (LLMs) like GPT-4~\citep{RLHF} and others, e.g. PaLM 2~\citep{anil2023palm} and LLaMA~\citep{touvron2023llama} have achieved a high degree of language fluency, enabling them to generate coherent and meaningful text-based responses. Moreover, the Transformer has excelled in pattern-matching tasks across various domains, including vision tasks like object detection and image segmentation~\citep{segmentAnything}. This advancement in deep neural networks (DNNs) may be expected due to the efficient use of global context; however, the more exciting development is evidence showing the architecture's ability to learn mechanisms for solving both familiar and new problems, rather than just pattern matching or memorizing.  Specifically, the models are starting to show the ability to synthesise complex algorithms via the composition of tools,  for example, Toolformer~\citep{toolformer} and Visual programming ~\citep{visual_program}. In addition to this, we are seeing RL-based decision-making agents also induce algorithms, for example, AlphaTensor~\citep{alphatensor} and AlphaDev~\citep{alphadev}.

Viewing learning as a hierarchical, discrete composition signals a paradigm shift in AI's future. Traditionally, deep networks learned specific algorithms, their parameters defined by weights adjusted via gradient descent. This method suits pattern recognition but falls short of general algorithm learning, as it doesn't facilitate the explicit reuse of learned skills or code for new tasks which is important to achieve scalable and sample-efficient learning~\citep{voyager}. In contrast, in recent works, we are seeing architectures with the capability to compose a complex algorithm from simpler algorithms. For example,~\citet{visual_program} tackles the task of identifying the key characters in an image, this task requires face detection, retrieving character names from a knowledge base and pairing names to faces. Learning this task end-to-end would be difficult, as it would require the system to learn each aspect from scratch, which is unnecessary because such models already exist and should not need to be relearned. The importance of learning these algorithms lies in the fact that solutions to problems are ultimately encoded as algorithms, built from the composition of sub-functions. Thus, a model that can construct algorithms represents significant progress towards a general model that can generate solutions for any task.

It is important to emphasise that many of the works which demonstrate the ability to compose using language models have some key limitations. Either the model is trained to compose external tools (eg. Toolformer~\citep{toolformer}) or the composition occurs within the weights of the model. In the first case some of the execution occurs outside the model and in the second case the execution is limited by the programs the architecture can easily execute.

We also argue that it is important for the architecture to be able to correctly and easily execute algorithms within the architecture (implicitly) since the model has explicit control over the sub-algorithms it is composing and can improve/modify their performance during the typical (e.g. gradient decent-based) training of the model. 

\begin{figure}[t]
  \centering

    \begin{subfigure}{0.45\textwidth}
    
  \includegraphics[width=1\linewidth]{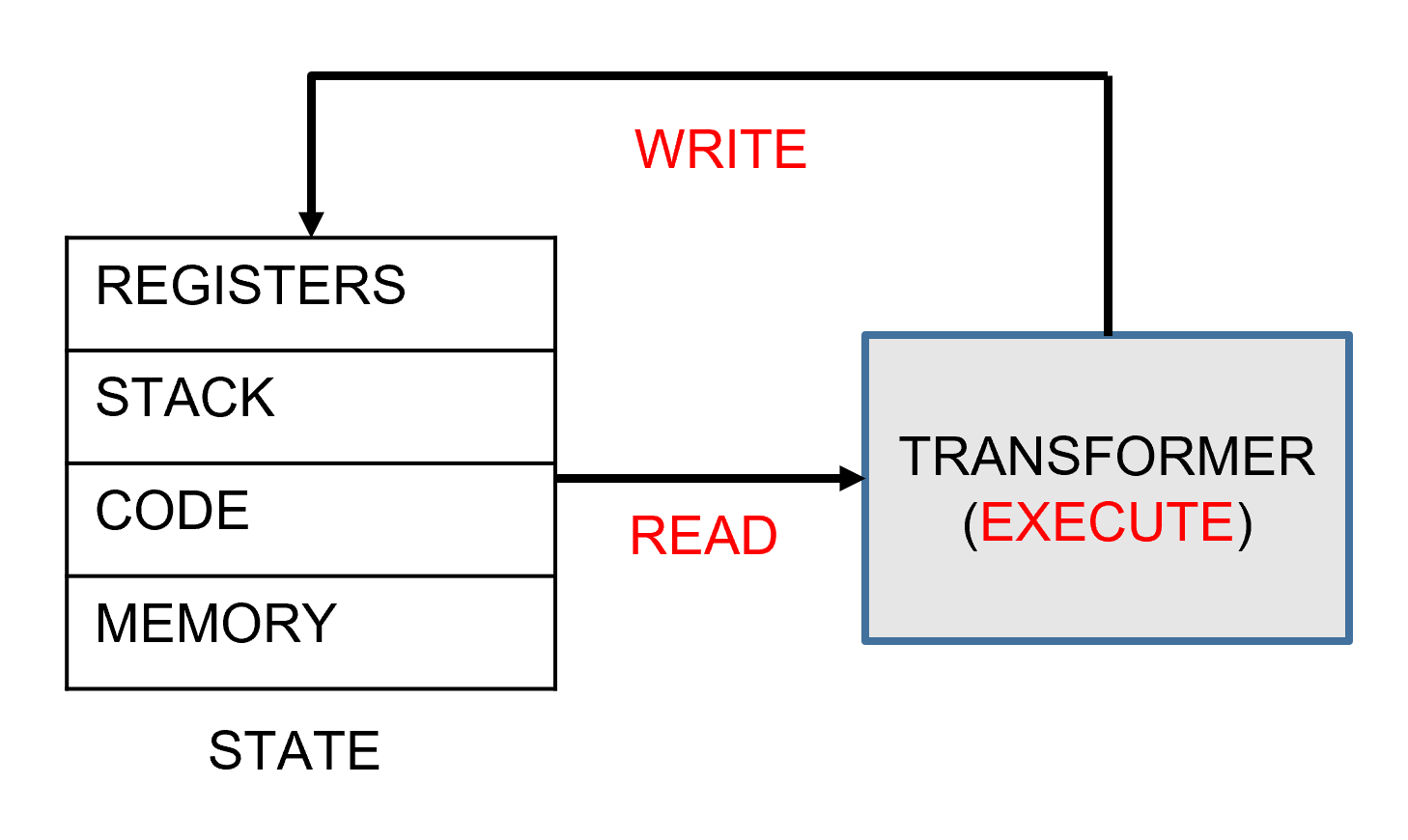}
  \caption{}
  \label{fig:mainmethod}
   \end{subfigure}       
    \begin{subfigure}{0.36\textwidth}
  \includegraphics[width=1.0\linewidth]{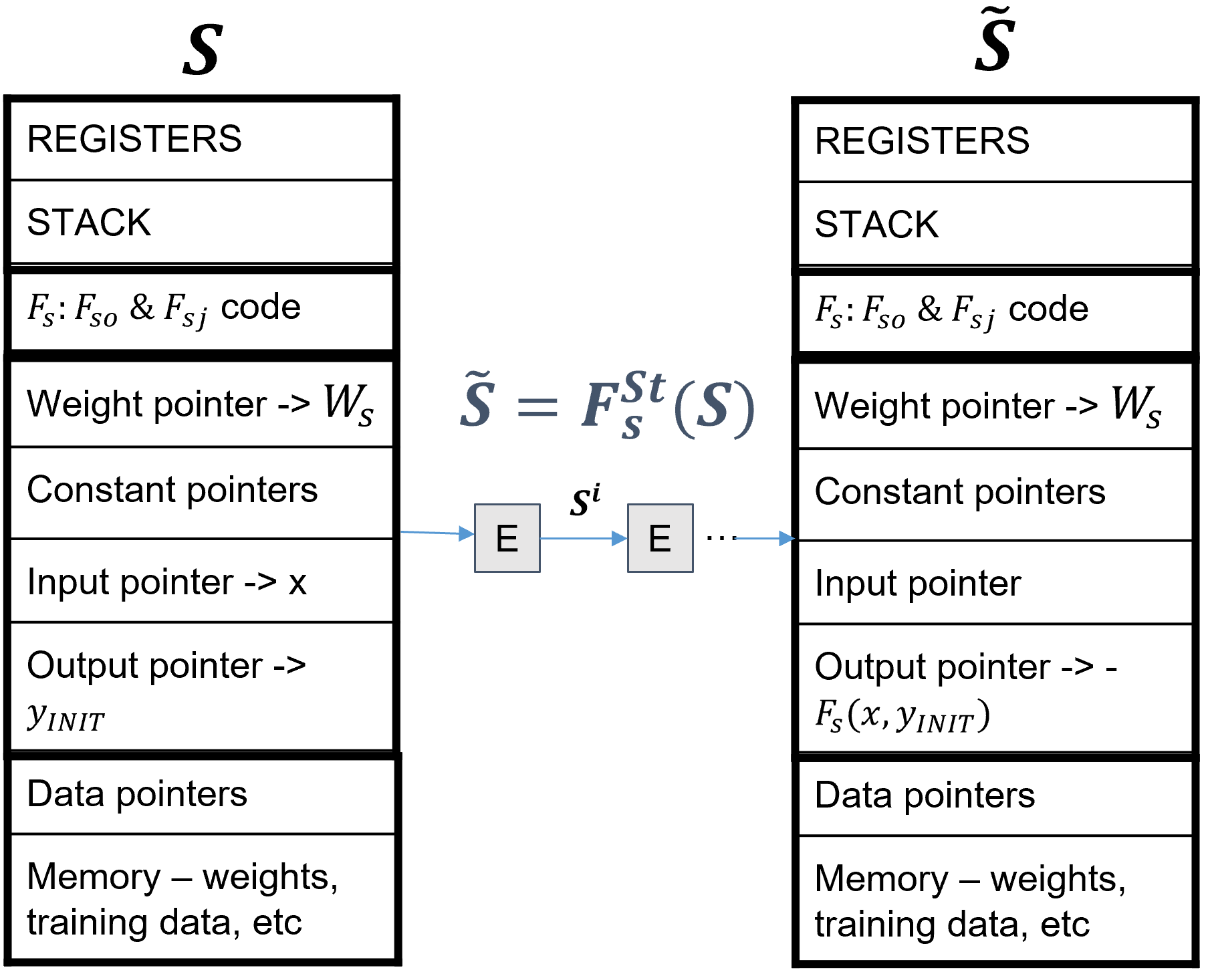}
  \caption{}
  \label{fig:input_struct}
  \end{subfigure}    
  \caption{Diagram in part (a) provides a simplified representation of our proposed architecture. The input is a state made up of blocks that store assembly code, memory for data and registers and stack for execution. The code stored in the state is then executed iteratively via a Transformer 'CPU' and the results are written back into the state which is outputted. Part (b) provides a minimal example of the core components that make up the state.}
\end{figure}
In this position paper, we explore the idea that developing a task-agnostic Transformer-like architecture which can learn to compose a solution for (nearly) any task may not be as difficult as one may think. We propose a theoretical framework of how it could be constructed. We use the terms skills, sub-functions and sub-algorithms interchangeably for different levels of abstraction for composition. Our key constraint in the design is that we want to learn to compose sub-algorithms learnt by our architecture implicitly as opposed to an explicit composition as done in many program synthesis works. In particular, we assume that all the learnt skills of our system are represented by our Transformer-like architecture. When a more complex skill is being learnt these skills are provided as input to our architecture. The key challenge that we address in this paper is that we propose how the execution and composition of such skills can be achieved as a simple forward pass through our Transformer-like architecture. 
It consists of the following key parts.\\ \noindent \textbf{Code of skills as an input.} First, unlike common feed-forward networks, our input consists not only of typical data input such as images but also of 'assembly' code and weights of the sub-skill networks (see Section \ref{architecture}). This is beneficial since in traditional networks to learn algorithms from scratch the smallest skills (even multiplication) have to be learnt. Hence, blocks of executable code are a rather efficient starting point as opposed to vanilla MHA ~\citep{transformer}, MLP~\citep{mlp}, LSTM~\citep{LSTM} mechanisms. \\
\noindent \textbf{State as input and output.} The input and output of the network both take the same form of a state which represents a computational model made up of registers, memory and blocks for skills which include code, data, inputs, intermediates and outputs for the executable skills.\\
\noindent \textbf{Simple, sublinear recursive 'CPU' Transformer layer.} 
The depth of the Transformer network is decoupled from the length of the program to execute.\\
\noindent \textbf{Learning a complex skill as a mixture.} We propose a mechanism for learning a complex skill via a recursive compositional mixture of previously learnt skills which in theory is capable of constructing any computer program given assembly instructions as base atomic skills to compose up from. Additionally, we propose two training schemes to learn the parameters of the mixture - one gradient-based and the other tree-search-based.\\
\noindent \textbf{Hierarchical curriculum learning.} To learn a hierarchical skill we propose a hierarchical curriculum learning approach. A hierarchical skill is learnt by iteratively training our network to learn a set of skills from a set of already learnt skills. Once a skill is learnt it is appended to the input to be used towards learning other skills.

The remainder of this work is structured as follows. We provide an overview of related work on algorithm learning in Section~\ref{related_work}. We then describe our proposed Transformer-like architecture for algorithm induction in Section~\ref{method} and provide final discussions in Section~\ref{discussion}.

\section{Algorithm Learning}
\label{related_work}
In this section, we will review the different types of approaches to algorithm learning, that is the the task of constructing an algorithm/program to solve a task. They can be split up into the following dimensions: i) induction vs synthesis. In the former group, the model learns an internal representation of the algorithm within the weights (induction), whereas in the latter group, the model outputs a representation of the algorithm e.g. code (synthesis). ii) Implicit vs explicit execution: the model executes the entire learnt algorithm within the network (implicit execution) vs some/all explicitly executed outside the model. iii) Task learning based on purely input-output pairs vs augmented with descriptions and other inputs.

\subsection{Algorithm induction from only input and output examples} \label{pure_io}

In this section, we explore the capability of DNNs to induce algorithms from just input-output examples for both specific and general algorithms.

\subsubsection{Specific algorithm induction} \label{specigic_algo_induc}
These works analyse specific examples of algorithm induction intending to understand how the Transformer architecture learns and represents algorithms to explain the capabilities we see from such models.\\
\textbf{Modular addition}: \citet{nanda2023progress} train a Transformer based model for modular addition and demonstrate a phenomenon known as "grokking" where the model undergoes a phase change and suddenly generalises to complete the task perfectly demonstrating that the model learns the correct algorithm to complete the task. In this case, reverse engineering the weights revealed the Transformer had learnt to compose discrete Fourier transforms and trigonometric identities to solve the task. However, this composition of a specific set of functions might not necessarily be sufficient for arbitrary algorithms.\\
\textbf{Simulating gradient decent} \citet{in-context} present another example, they provide evidence that Transformer-based models are implementing standard learning algorithms implicitly and constructing small models via their activations and training these models, they prove Transformers can simulate gradient descent to learn linear models to construct new predictors when provided a sequence of labelled examples, $(x,f(x))$. Here the model has effectively learnt to compose algorithms to create models and train the weights on the model to build a classifier. This is interesting as it is clear that for some softer pattern-matching tasks such as classification, constructing and training a deep network is the best algorithm to solve the task and is something the Transformer can implicitly execute to some extent. The model we propose is also capable of executing and composing such sub-networks.

These examples demonstrate that the sophisticated capabilities of these models may arise via the construction, composition and execution of algorithms towards correctly solving a task. This is very promising as this is fundamentally how a solution to any general problem must be formed. While these are examples of specific algorithms that have been induced we would like to understand the capability of inducing arbitrary algorithms.

\subsubsection{General algorithm induction} \label{general_algo_induc}

The earliest modern deep learning era works looking at algorithm induction included Learning to Execute~\citep{learning-to-execute} which looked at feeding simple programs as input to an RNN and the model aims to predict the output of the program, if it can do so correctly it will have learnt to execute the program. They were able to train to add two 9-digit numbers with 99\% accuracy, however, it remained difficult to conclude if the model had learnt to construct a correct addition algorithm or not. Our proposed skill composition parameters allow inspection of the parameters to reconstruct the algorithm and check for correctness. Neural Turing Machines (NTM)~\citep{neural-turing} and Differential Neural Computers (DNC)~\citep{DNC} are other early works, they look at augmenting DNNs with memory as this is an important component of program execution. NTMs were able to infer simple programs such as copying and sorting and DNCs were able to construct some algorithms over graph-structured data. These were also RNN-based and it is unclear how they scale to construct algorithms for more complex tasks. 

The Transformer architecture on the other hand has been demonstrating remarkable capabilities in a range of domains demonstrating the flexibility in the types of algorithms it can construct. There have been some works that aim to explore which algorithms can be represented by the Transformer architecture with specific weights. TACR~\cite{tracr} and RASP~\cite{RASP} show how some limited sets of programs (e.g. computing token frequencies, sorting, and parenthesis checking) can be represented and most importantly which, in theory, could be easily learnt by Transformer architecture. While showing representations of programs in the weights is promising the important question is if the programs can be learnt from input-output examples. Firstly, how do we evaluate if a model has learnt an algorithm correctly? Length generalisation (the ability of a model to successfully generalise to inputs longer than those found in the training data) is a strong indicator that the model has learnt the correct algorithm for the task as opposed to memorising the training data. For example, testing a model to add numbers with more digits than those in the training data. \citet{length-generalization} provide evidence towards the "RASP-Generalisation conjecture", which states that Transformers tend to length generalize on
a task if the task can be solved by a short RASP program which works for all input lengths.

While these papers adhere to the vanilla Transformer architecture they are also limited by it. The Transformer may be ultimately limited to simple programs as composing anything more complex is hard via gradient descent as the composition is discrete. Additionally, the vanilla Transformer architecture cannot compute programs to arbitrary recursion depth due to the fixed computation and this may pose an additional challenge to learning more complex algorithms, Looped Transformers~\citep{looped} are a way to address this, more details in Section \ref{implicit_exec}. Another important point is that pure input-output training itself is limited since the information content of the training set may not be sufficient to learn the correct algorithm. As humans, we solve problems by drawing on all previous knowledge not just on the current examples provided.
\subsection{Algorithm induction from rich input}
We now consider methods that aim to construct algorithms but also have access to additional information such as descriptions or pre-trained weights (e.g. LLMs) to supplement the input-output examples. There are two forms in which the constructed algorithm may be executed, either all execution happens within the network (implicit) or some/all of the execution happens externally (explicit). We look at both types that use rich input-output.

\subsubsection{implicit execution} \label{implicit_exec}
Transformer-based language models, such as GPT4, have shown remarkable few shot learning capabilities via in-context learning and hence demonstrate an additional form of implicit execution beyond the model storing the algorithms within the weights of the network that are then implicitly executed. In-context learning shows the model can interpret and execute instructions/code in the input.\\
\textbf{Transformers as interpreters.} Through the layers of the model, the Transformer is capable of mimicking the execution of a subset of general algorithms. It is possible to prompt ChatGPT in such a way as to emulate a virtual machine or a Python interpreter~\citep{chat_vm}. ChatGPT is then able to simulate the execution of programs, hinting at a similar simulation when attempting to do algorithmic tasks. The execution of programs within the simulated ChatGPT Python interpreter does not work perfectly as it is not 'real' execution. The same problem persists when attempting to complete simple algorithmic tasks such as multiplication. To produce the outputs of the execution it has learnt to compose the simulated execution of the lines of code written in the prompt. However, since the model does not actually explicitly execute these lines and hasn't properly induced those functions the execution isn't correct. Our method on the other hand ensures correct execution of assembly commands to make sure this is not an issue.\\
\textbf{In-context learning.} LLMs improve at a task if provided examples and guidance in context i.e. in the input, without any modifications to the weights of the model, a phenomenon known as in-context learning~\citep{in-context}. Importantly, via in-context learning the model can solve a task it had never seen in the training data. This suggests that the model has learnt to compress the task of being shown examples and completing a task in a way that aligns with the examples to an algorithm as opposed to relying on statistics for what comes next. \cite{olsson2022context} propose that in-context learning can be explained by induction heads. 
They propose that 'this might constitute the mechanism for the actual majority of all in-context learning in large Transformer models'. The key point is that the Transformer model has learnt some procedures to replicate patterns in the input to complete unseen tasks instead of producing outputs purely based on the statistics of the training data. Note, the in-context learning algorithm can be used for natural language instruction execution, i.e. executing more general algorithms. \citet{algo-in-context} show that the algorithmic reasoning capabilities of LLMs are improved by teaching algorithmic reasoning in context. This shows that the Transformer effectively uses in-context learning to follow a procedure provided in the prompt which can be viewed as 'execution' of natural language. In our model, we take this to the extreme by providing code in the input that can be executed explicitly by our model towards solving the task.\\
\noindent \textbf{Looped execution.} \cite{looped} take an approach to build a Transformer-based architecture capable of universal computation, a Looped Transformer. They start with executing programs written in a single instruction set language SUBLEQ \cite{SUBLEQ}, which is shown to be capable of defining a universal computer. They construct a 9-layer Transformer capable of executing the SUBLEQ command which requires the Transformer to be able to subtract and conditionally branch to execute each instruction. The program is stored in the input and can be executed by running the Transformer layer in a loop. This allows additional code and functions to be provided in the input to be made use of. This is close to what we propose but this work just looks at how to execute code with a Transformer layer, we take it further to propose how to learn programs with such an architecture (Section \ref{composing_an_algo}). 

\subsubsection{explicit execution}
The additional information supplied to the model can be in the form of external tools. This has been a popular approach recently with GPT4 able to make function/API calls and \citet{toolformer} make use of this approach for Toolformer. 
This approach addresses the challenges LLMs face with basic arithmetic, which often impedes their ability to learn tasks requiring such calculations due to the inherent difficulty in carrying out arithmetic operations. Tools get around this by allowing for correct arithmetic that can be called upon without learning those algorithms. While this is promising, this requires execution outside of the model (explicit execution) which is not ideal as the model is unable to modify these algorithms. This is important as some algorithms may need to be tweaked/adapted in certain situations. 

\subsection{Program Synthesis}
In this section, we look at an alternative way to learn algorithms to solve a task. Program synthesis includes methods that generate code or some representation of the algorithm instead of the solution. 
\subsubsection{Transformer-based}
Examples of synthesis with Transformers include AlphaCode~\citep{alphacode} and Codex~\citep{codex}, these are language models that are trained to produce computer programs from problem descriptions. While these methods have been quite capable with AlphaCode2 boasting a performance in the top 15\% of competitive programmers there is a fundamental limitation. LLMs suffer from hallucinations and to make sure the code is robust to this there needs to be some level of verification in a feedback loop, if the model is unable to execute the code correctly internally it cannot do this reliably. Another example is Boolformer~\citep{boolformer} which takes truth tables and converts them to their logical functions. 

FunSearch~\citep{FunSearch} addresses this limitation by pairing a language model with an evaluator, similar to FCPS~\citep{fcps}, which evaluates the synthesised algorithm and runs this in a feedback loop. This is a promising approach but can be taken one step further as we do in our proposed architecture, if the model itself is capable of correct arbitrary execution inside the network this feedback evaluation process can be done internally and the network can improve and modify the meta-evaluation-feedback improvement algorithm as needed.

\subsubsection{RL-based}
AlphaDev~\citep{alphadev} is an RL-based agent that composes assembly commands to synthesise programs. Composing assembly commands brings the approach closer to learning general algorithms. Despite managing to learn an optimisation for a relatively simple algorithm the combinatorial complexity of composing assembly commands still proves to be a challenge for AlphaDev towards learning more complex algorithms. Additionally, the training data comprises pure input-output the disadvantages of which were discussed in Section \ref{pure_io}. This is similar to the approach we take as we also have base-level commands as some form of assembly, our approach however can leverage language models to inform the composition to prune the search and employs composing in a hierarchy to make composing long programs more tractable.

\section{Method}
\label{method}

In this section, we propose (i) a formalisation of the learning task of skills in Section~\ref{task}, (ii) a network architecture capable of executing and combining trained networks in Section~\ref{architecture} and (iii) a scheme in which to train to solve a collection of tasks in a hierarchical fashion in Section~\ref{hierarchy}. 

\subsection{Learning task}
\label{task}
 The goal of an architecture capable of explicit execution of trained skills and algorithmic composition of these skills can be mathematically formulated as follows. We have a set of trained skills $\mathcal{F} = \{ F_{1}, F_{2},...,F_{n}\}$, and a set of parameters $\mathcal{W} = \{W_{1}, W_{2},...,W_{n}\}$. In the most general sense each skill $F_{k}$ is just a function parameterised in some way by the corresponding set of weights $W_{k}$ such that the function takes an input $\mathbf{X}$ and produces the mapping, $F_{k}: \mathbb{R}^{D_x} \times \mathbb{R}^{D_w} \rightarrow \mathbb{R}^{D_y}$, $Y = F(\mathbf{X}, W_k)$. When presented with a new task with inputs $\mathbf{X}$ and labels $\mathbf{y}$ a function to solve the problem, $F_{C}$, is constructed in the following way: $\mathbf{y_{pred}} = F_{C}(\mathcal{F}, \mathcal{W}, \mathbf{X}) = G(\mathcal{F}, \mathcal{W})[\mathbf{X}]$. $F_{C}$ crucially takes the skill networks themselves as inputs and composes these functions in some way specified by $G$ and this is then applied to the task inputs $\mathbf{X}$. This formulation frames solving the new task by composing previous skills, theoretically allowing for the formation of algorithmic programs composed of sub-functions.

 \subsection{Network Architecture}
 \label{architecture}
 The key components include a change to the input structure to include networks themselves as inputs and programmatic execution of these networks as sub-skills for a complex skill. 

\subsubsection{Input Structure} 

\begin{figure}[t]
  \centering
 
  \begin{subfigure}{0.45\textwidth} 
  \includegraphics[width=1.0\linewidth]{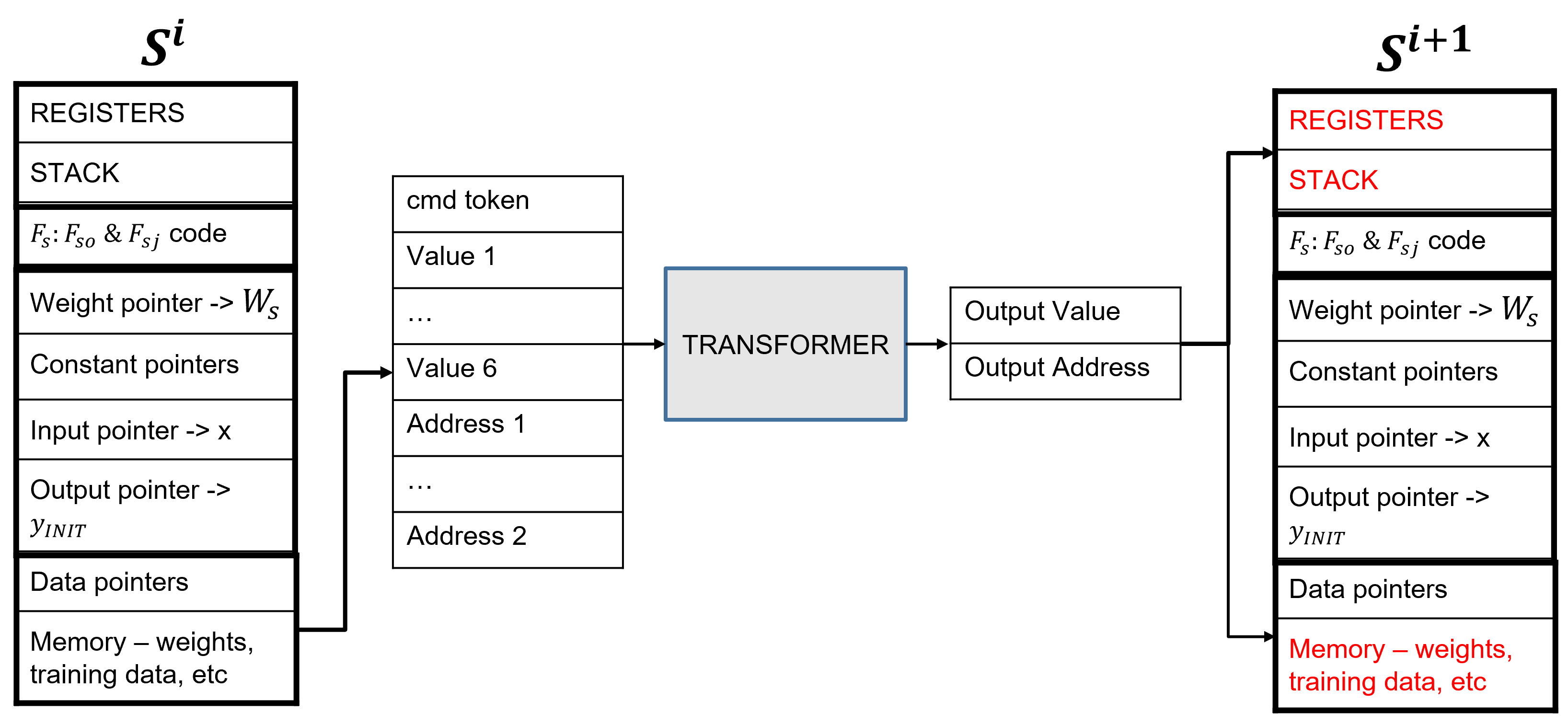}
  \caption{}
  \label{fig:exec_diagram}
   \end{subfigure}
   \begin{subfigure}{0.4\textwidth} 
       \includegraphics[width=1.0\linewidth]{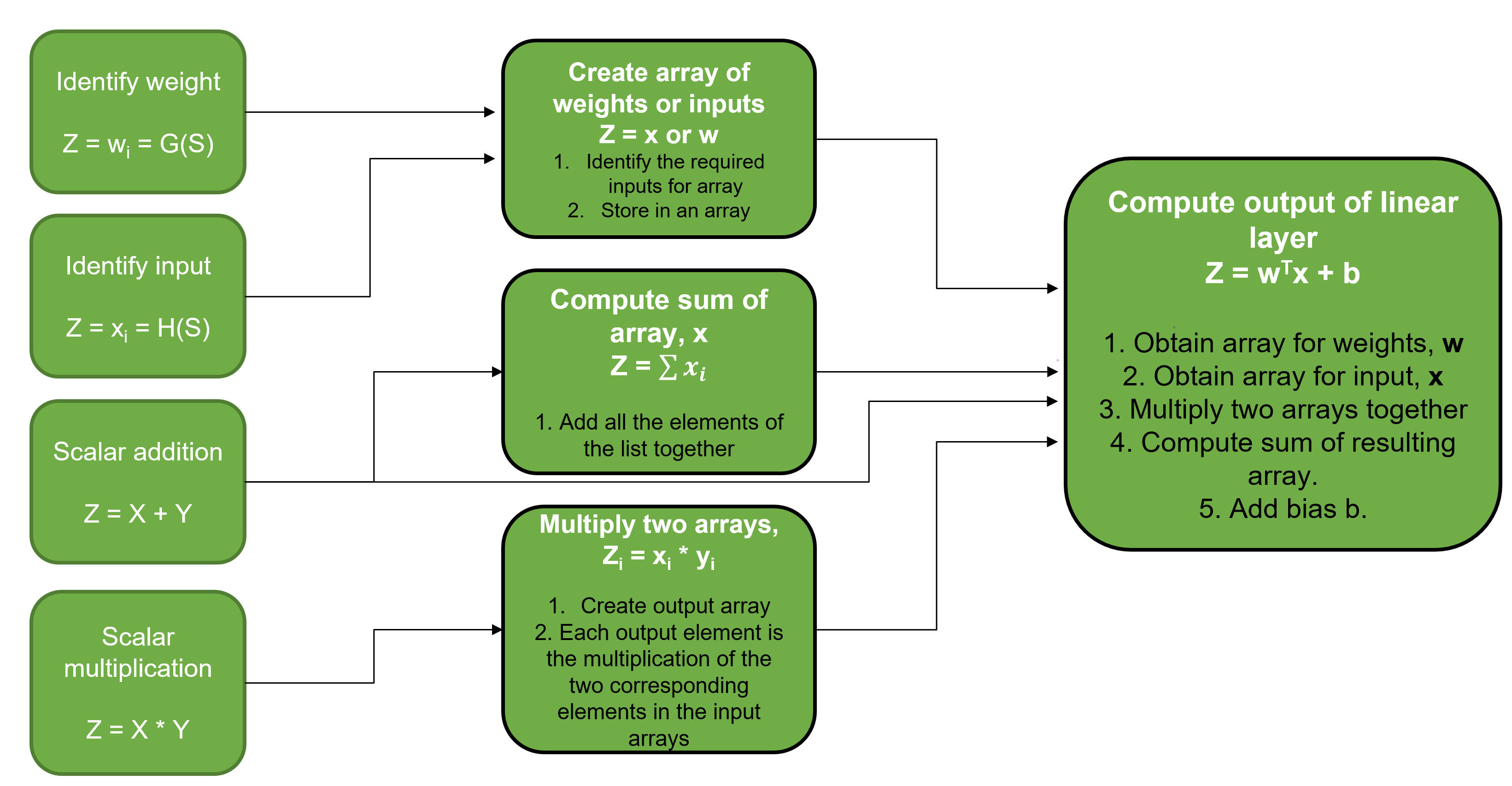}
  \caption{}
  \label{fig:hier}
   \end{subfigure}
  \caption{Part (a) describes the execution of a single assembly command, $ S^{i+1} = E(S^{i})$, S is the state and E is the Transformer execution block (more details in Figure \ref{fig:input_struct}) Part (b) describes the hierarchy of skills to compute the linear layer, $Z = w^{T} + b$.} \vspace{-0.4cm}
\end{figure}

For the proposed architecture we move away from a traditional network that maps inputs (e.g. raw pixels of an image) to outputs (e.g. probability distribution over classes), $Y = F_s(X,W_s)$, and instead, we formulate learning as an iterative update of an input made up of more than just traditional inputs which we refer to as the state, $ S \in \mathbb{R}^n$. An update to the state can be made by executing the code for the network $F_{s}$ stored in the state. This update maps $S$ to $\tilde{S}$ in the following way: $\tilde{S} = F_s^{St}(S), F^{St}: \mathbb{R}^n \rightarrow \mathbb{R}^n$. $F_{s}^{St}$ denotes executing the code for a skill $s$ stored in the state. Figure \ref{fig:input_struct} gives a minimal representation of the input with just one skill block. At a high level, the state is made up of a few key blocks.

\noindent \textbf{Skill blocks.} For each sub-skill block there is code and slots for pointers to relevant constants, inputs, outputs and weights for the skill. The pointers point to locations in the static memory where this data is actually stored. 
    The code stored for a skill $F_s$ is made up of two parts: 
    $F_s = F_{so}$ (if $M_{s} = 1$) and $F_s=F_{sj}$ (if $M_{s} = 0$), 
    where $F_{so}$ computes the output for the network, $F_{sj}$ computes the Jacobian with respect to the weights and $M_s$ is a boolean stored in the state which can be used to choose which of these to compute.\\
\noindent \textbf{Static memory.} The static memory stores any pre-existing inputs to the skills such as labelled training data and weights for networks and the memory has free space for storing any outputs resulting from the execution of the skills.\\
\noindent \textbf{Data pointers.} At the top of the memory a list of pointers is stored, referred to as data pointers; this is a list of pointers to all pieces of data stored in the memory. These data pointers can then be loaded into functions to pass by reference desired inputs to a skill.\\
\noindent \textbf{Registers and stack.}  Since the subskills are stored and executed via assembly, this requires blocks for registers and a stack. The execution of the assembly code builds a stack when creating local variables. The registers are used to hold intermediate values when executing assembly commands. An important register is $R_L$, which is the line register which points to the line in the assembly program to execute.

\subsubsection{Programmatic Execution} 

Execution of the code for the network equates to executing the sequence of assembly commands represented by the repeated use of the execution module E shown in Figure \ref{fig:input_struct}, giving $\tilde{S} = E(E(...E(S)...))$ and the intermediate states are denoted, $ S^{i} = E(S^{i-1})$. The execution module determines the location of the command to execute from the $R_L$ register, then reads the values for the inputs and outputs from the addresses, executes the command specified by the token via a Transformer and takes the resulting output value and stores it in the output address in the state, as shown in Figure \ref{fig:exec_diagram}. Note, Figure \ref{fig:input_struct} can also be used to understand the execution of trained skills sequentially as part of a complex skill. The parallel here is that each assembly command can be treated as a skill or sub-function of its own and these are composed to build up the main function. 
\subsubsection{Composing an algorithm} \label{composing_an_algo}
Next, we propose a method to allow the network to explicitly execute a learnt skill stored in the state and combine skills towards solving a new more complex task. We denote this new complex skill as $F_C$. We can formulate this new skill as the recursive application of a mixture of skills.

At the top level the skill composition skill, also stored in the input, $F_C^{St}(S^{i}, t)$ is recursive in nature; for each level of the recursion, the state is updated from $S^{i}$ to $S^{i+1}$ and then passed back into the skill composition function this time with the $t$ parameter decreased by one as shown in Equation \ref{eq:1}. 
\begin{align} \label{eq:1}
F_C^{St}(S^{i}, t) &=
    \left \{
	\begin{array}{ll}
		S^{i+1}  & \mbox{if } t = 0 \\
		F_{C}^{St}(S^{i+1}, t-1) &  \text{otherwise}
	\end{array} 
\right. 
\end{align}
At each recursion level, the state is updated according to Equation \ref{eq:2}. This is a weighted sum of the resulting state after executing the available skills denoted by the set $S$.
\begin{align}\label{eq:2}
    S^{i+1} &=  \sum_{s \in S} F_{comp}(S^{i}, s, t, W^{\alpha}) 
\end{align}
Equation \ref{eq:3} shows how the result of each skill is computed. First, the $W^{in}$ function is used to compute which input, output and data pointers should be copied into the relevant cells for the skills. This selects which inputs, output locations and data are processed by the skill. $W^{in}$ returns a sparse matrix which is multiplied by the state to load the relevant data pointers to the desired places. The skill is then executed via the state, $F^{St}(s)$. The result is then weighted by $\alpha$ which controls which skills are used at each step in the algorithm. 
\begin{align} \label{eq:3}
    F_{comp}(S^{i},s,t,W^{\alpha}) &= \alpha F^{St}(s)(W^{in}(s, S^{i},t)S^{i}) 
\end{align}
In general, $\alpha$ is computed via an arbitrary function $F_{\alpha}$ and is a function of the state, skill, timestep, any prior initialisation of the parameter provided $\alpha_{0}$ and perhaps if $F_{\alpha} $ is a deep network, any parameters of the network $W_{\alpha}$, shown in Equation \ref{eq:4}. We propose a language model to guide the training and produce the $\alpha$s, the language model is now the function $F_{\alpha}$ and $W_{\alpha}$ correspond to the weights of the language model. The language model takes in the current state and timestep to propose a sophisticated choice of $\alpha$.
\begin{align} \label{eq:4}
    \alpha &= F_{\alpha}(W_{\alpha},s, S^{i}, t, \alpha_{0})  
\end{align}
Note the mixture of the skills here can be viewed as a generalisation of multi-head attention with sophisticated attention weights.

\textbf{Gradient-based learning.} $F_{C}$ can be trained over the parameters for the functions $W^{in}$ and $F_{\alpha}$, which control the pointers loaded into the functions for inputs and outputs and the sub-skills executed at each time step, allowing for the execution of sub-skills towards solving the main task. \\
\textbf{Tree-search-based learning.} Since the composition is discrete, other learning algorithms that operate on discrete parameters are preferable. For example, Monte Carlo tree search is a vital component of the recent successes in discovering algorithms such as AlphaTensor ~\citep{alphatensor} via AlphaZero~\citep{alphazero} and could be incorporated towards general algorithm learning. Due to the large combinatorial complexity, we propose pruning the search with a language model. This formulation in theory has the flexibility to compose any program. Every program has a deterministic output for a given input. The input to the function is stored in the state $S$; all the parameters that dictate the composition of learnt skills are a function of this and hence have the flexibility to compose skills in all possible ways over the space of all possible inputs.

\subsection{Hierarchy of Skills}
\label{hierarchy}
Next, we describe a framework to build up the ability of the agent to complete increasingly complex tasks via a hierarchy of skills. The hypothesis is that every complex skill can be broken down into sub-skills that can be composed, and learning the composition via a hierarchy is an efficient and scalable way to learn. A collection of skills can be set up with test data for each skill. We start with trained atomic skills, then at the first generation of skill learning any skills that can be learnt by composing the atomic skills are solved, testing each against the test data. These newly solved skills are now added to the collection of trained skills. Then in the following generation new skills are solved continuing in a hierarchical fashion, reminiscent of human learning. This process for computing a linear layer is shown in Figure \ref{fig:hier}. 

In theory, this should be able to compose algorithms but the combinatorial complexity could pose a challenge for training. We can provide text descriptions of how to complete more complex tasks, analogous to how a textbook may teach. Given descriptions of the tasks, the language model can generate a sequence of predictions for the distribution over the $\alpha^{t}_{s}$ parameters to indicate which sub-skills are most relevant at a given time step. This can form the initialisation of the model allowing for faster learning of the correct composition of sub-skills.

\section{Potential impact and limitations of our approach} \label{discussion}
We speculate some benefits of our proposed approach below: \\
\noindent \textbf{Merging "hard" and "soft" algorithms.} This approach allows for the learning of features from data by whatever processing is necessary followed by precise algorithms learnt for processing these features. An example of this could be features extracted for two images and these being processed by the RANSAC algorithm for image alignment.\\
\noindent \textbf{Reuse feature extractors.} This approach allows the model to have the flexibility of using sub-networks created for other tasks that produce useful features that would be useful for the current task, as opposed to ensemble methods which are only able to make use of the outputs of other trained networks.~\citet{feature-reuse} show that reusing convolutional features can be successfully applied to training image classifiers with limited data.\\
\noindent \textbf{Eliminate catastrophic forgetting.} Since previously learnt skills are not modified and simply combined, this makes the method robust to catastrophic forgetting, though we assume we have an 'infinite' capacity to store information and code for learnt sub-skills. This also means that, unlike fine-tuning, if the model tries to learn a skill it is not able to learn yet since it lacks the correct sub-skills required, this will not affect the model's ability for any already learnt skills. Voyager~\citep{voyager}, leverages a similar approach to show the elimination of catastrophic forgetting. \\
\noindent \textbf{Precise skill execution.} Our approach emphasizes the explicit execution of learned skills, allowing for precise skill composition crucial for creating correct algorithms. While Turing-complete networks might theoretically learn to execute sub-algorithms, the complexity of even basic algorithms like multiplication, which require numerous steps, poses a challenge for learning composition and execution. Direct access to an algorithm for execution as a whole simplifies skill utilization, facilitating the construction of correct algorithms. Toolformer~\citep{toolformer} illustrates the benefits of providing Transformers with executable tools or skills, enhancing their ability to tackle more complex tasks. By integrating skills directly, we enable the Transformer to bypass the initial learning phase, streamlining the process towards mastering complex skills. This strategy proves especially effective in specific domains where certain problem aspects are already well-addressed, eliminating the need for the model to learn these sub-functions from scratch.\\
\noindent \textbf{Interpretability.} The layer we propose to learn to combine skills also brings interpretability benefits which become increasingly important as we start to deploy AI systems in critical situations, the skills used by the model to construct a solution to the task can be inspected retrospectively to understand the approach model used to solve the problem. 

We recognise some limitations of our proposed methodology:\\
\textbf{Computational efficiency.} There are some challenges associated with the computational efficiency of the proposed computations as our framework scales. If there are many sub-functions in the input the network that runs to pick the $\alpha$'s smartly would require a lot of computation.\\ 
\textbf{Learning hierarchical skills.} The hierarchies for the hierarchical learning approach are not always straightforward to construct. The language model will attempt to propose hierarchies and try to solve the problems within the hierarchies recursively and adapt the hierarchies as needed, it is unclear how well these models will be able to tackle this.  
\section{Conclusion}
\label{conclusion}

We propose that decision-making for task-solving is essentially a sequential algorithm induction problem,
reducing the decision-making process for any problem to the low-level choice of sequential instructions in the algorithm. We explore the current methods and architectures for algorithm learning and note weaknesses across the existing classes of solutions. We propose a Transformer-like architecture which allows for specific and arbitrary implicit execution, a parameterization for learning composition and a learning framework to hierarchically construct increasingly complex algorithms. We hope that our work will contribute to the ML communities' search for DNN architectures and learning methods which are task-agnostic.

\bibliography{iclr2024_conference}

\begin{thebibliography}{32}
\providecommand{\natexlab}[1]{#1}
\providecommand{\url}[1]{\texttt{#1}}
\expandafter\ifx\csname urlstyle\endcsname\relax
  \providecommand{\doi}[1]{doi: #1}\else
  \providecommand{\doi}{doi: \begingroup \urlstyle{rm}\Url}\fi

\bibitem[Akyürek et~al.(2022)Akyürek, Schuurmans, Andreas, Ma, and Zhou]{in-context}
Ekin Akyürek, Dale Schuurmans, Jacob Andreas, Tengyu Ma, and Denny Zhou.
\newblock What learning algorithm is in-context learning? investigations with linear models, 2022.
\newblock URL \url{https://arxiv.org/abs/2211.15661}.

\bibitem[Anil et~al.(2023)Anil, Dai, Firat, Johnson, Lepikhin, Passos, Shakeri, Taropa, Bailey, Chen, Chu, Clark, Shafey, Huang, Meier-Hellstern, Mishra, Moreira, Omernick, Robinson, Ruder, Tay, Xiao, Xu, Zhang, Abrego, Ahn, Austin, Barham, Botha, Bradbury, Brahma, Brooks, Catasta, Cheng, Cherry, Choquette-Choo, Chowdhery, Crepy, Dave, Dehghani, Dev, Devlin, Díaz, Du, Dyer, Feinberg, Feng, Fienber, Freitag, Garcia, Gehrmann, Gonzalez, Gur-Ari, Hand, Hashemi, Hou, Howland, Hu, Hui, Hurwitz, Isard, Ittycheriah, Jagielski, Jia, Kenealy, Krikun, Kudugunta, Lan, Lee, Lee, Li, Li, Li, Li, Li, Lim, Lin, Liu, Liu, Maggioni, Mahendru, Maynez, Misra, Moussalem, Nado, Nham, Ni, Nystrom, Parrish, Pellat, Polacek, Polozov, Pope, Qiao, Reif, Richter, Riley, Ros, Roy, Saeta, Samuel, Shelby, Slone, Smilkov, So, Sohn, Tokumine, Valter, Vasudevan, Vodrahalli, Wang, Wang, Wang, Wang, Wieting, Wu, Xu, Xu, Xue, Yin, Yu, Zhang, Zheng, Zheng, Zhou, Zhou, Petrov, and Wu]{anil2023palm}
Rohan Anil, Andrew~M. Dai, Orhan Firat, Melvin Johnson, Dmitry Lepikhin, Alexandre Passos, Siamak Shakeri, Emanuel Taropa, Paige Bailey, Zhifeng Chen, Eric Chu, Jonathan~H. Clark, Laurent~El Shafey, Yanping Huang, Kathy Meier-Hellstern, Gaurav Mishra, Erica Moreira, Mark Omernick, Kevin Robinson, Sebastian Ruder, Yi~Tay, Kefan Xiao, Yuanzhong Xu, Yujing Zhang, Gustavo~Hernandez Abrego, Junwhan Ahn, Jacob Austin, Paul Barham, Jan Botha, James Bradbury, Siddhartha Brahma, Kevin Brooks, Michele Catasta, Yong Cheng, Colin Cherry, Christopher~A. Choquette-Choo, Aakanksha Chowdhery, Clément Crepy, Shachi Dave, Mostafa Dehghani, Sunipa Dev, Jacob Devlin, Mark Díaz, Nan Du, Ethan Dyer, Vlad Feinberg, Fangxiaoyu Feng, Vlad Fienber, Markus Freitag, Xavier Garcia, Sebastian Gehrmann, Lucas Gonzalez, Guy Gur-Ari, Steven Hand, Hadi Hashemi, Le~Hou, Joshua Howland, Andrea Hu, Jeffrey Hui, Jeremy Hurwitz, Michael Isard, Abe Ittycheriah, Matthew Jagielski, Wenhao Jia, Kathleen Kenealy, Maxim Krikun, Sneha Kudugunta, Chang
  Lan, Katherine Lee, Benjamin Lee, Eric Li, Music Li, Wei Li, YaGuang Li, Jian Li, Hyeontaek Lim, Hanzhao Lin, Zhongtao Liu, Frederick Liu, Marcello Maggioni, Aroma Mahendru, Joshua Maynez, Vedant Misra, Maysam Moussalem, Zachary Nado, John Nham, Eric Ni, Andrew Nystrom, Alicia Parrish, Marie Pellat, Martin Polacek, Alex Polozov, Reiner Pope, Siyuan Qiao, Emily Reif, Bryan Richter, Parker Riley, Alex~Castro Ros, Aurko Roy, Brennan Saeta, Rajkumar Samuel, Renee Shelby, Ambrose Slone, Daniel Smilkov, David~R. So, Daniel Sohn, Simon Tokumine, Dasha Valter, Vijay Vasudevan, Kiran Vodrahalli, Xuezhi Wang, Pidong Wang, Zirui Wang, Tao Wang, John Wieting, Yuhuai Wu, Kelvin Xu, Yunhan Xu, Linting Xue, Pengcheng Yin, Jiahui Yu, Qiao Zhang, Steven Zheng, Ce~Zheng, Weikang Zhou, Denny Zhou, Slav Petrov, and Yonghui Wu.
\newblock Palm 2 technical report, 2023.

\bibitem[Chen et~al.(2021)Chen, Tworek, Jun, Yuan, Pinto, Kaplan, Edwards, Burda, Joseph, Brockman, Ray, Puri, Krueger, Petrov, Khlaaf, Sastry, Mishkin, Chan, Gray, Ryder, Pavlov, Power, Kaiser, Bavarian, Winter, Tillet, Such, Cummings, Plappert, Chantzis, Barnes, Herbert-Voss, Guss, Nichol, Paino, Tezak, Tang, Babuschkin, Balaji, Jain, Saunders, Hesse, Carr, Leike, Achiam, Misra, Morikawa, Radford, Knight, Brundage, Murati, Mayer, Welinder, McGrew, Amodei, McCandlish, Sutskever, and Zaremba]{codex}
Mark Chen, Jerry Tworek, Heewoo Jun, Qiming Yuan, Henrique Ponde de~Oliveira Pinto, Jared Kaplan, Harri Edwards, Yuri Burda, Nicholas Joseph, Greg Brockman, Alex Ray, Raul Puri, Gretchen Krueger, Michael Petrov, Heidy Khlaaf, Girish Sastry, Pamela Mishkin, Brooke Chan, Scott Gray, Nick Ryder, Mikhail Pavlov, Alethea Power, Lukasz Kaiser, Mohammad Bavarian, Clemens Winter, Philippe Tillet, Felipe~Petroski Such, Dave Cummings, Matthias Plappert, Fotios Chantzis, Elizabeth Barnes, Ariel Herbert-Voss, William~Hebgen Guss, Alex Nichol, Alex Paino, Nikolas Tezak, Jie Tang, Igor Babuschkin, Suchir Balaji, Shantanu Jain, William Saunders, Christopher Hesse, Andrew~N. Carr, Jan Leike, Josh Achiam, Vedant Misra, Evan Morikawa, Alec Radford, Matthew Knight, Miles Brundage, Mira Murati, Katie Mayer, Peter Welinder, Bob McGrew, Dario Amodei, Sam McCandlish, Ilya Sutskever, and Wojciech Zaremba.
\newblock Evaluating large language models trained on code, 2021.
\newblock URL \url{https://arxiv.org/abs/2107.03374}.

\bibitem[d'Ascoli et~al.(2023)d'Ascoli, Bengio, Susskind, and Abbé]{boolformer}
Stéphane d'Ascoli, Samy Bengio, Josh Susskind, and Emmanuel Abbé.
\newblock Boolformer: Symbolic regression of logic functions with transformers, 2023.

\bibitem[Degrave(2022)]{chat_vm}
Jonas Degrave.
\newblock Building a virtual machine inside chatgpt, Dec 2022.
\newblock URL \url{https://www.engraved.blog/building-a-virtual-machine-inside/}.

\bibitem[Fawzi et~al.(2022)Fawzi, Balog, Huang, Hubert, Romera-Paredes, Barekatain, Novikov, R.~Ruiz, Schrittwieser, Swirszcz, Silver, Hassabis, and Kohli]{alphatensor}
Alhussein Fawzi, Matej Balog, Aja Huang, Thomas Hubert, Bernardino Romera-Paredes, Mohammadamin Barekatain, Alexander Novikov, Francisco~J. R.~Ruiz, Julian Schrittwieser, Grzegorz Swirszcz, David Silver, Demis Hassabis, and Pushmeet Kohli.
\newblock Discovering faster matrix multiplication algorithms with reinforcement learning.
\newblock \emph{Nature}, 610\penalty0 (7930):\penalty0 47--53, Oct 2022.
\newblock ISSN 1476-4687.
\newblock \doi{10.1038/s41586-022-05172-4}.
\newblock URL \url{https://doi.org/10.1038/s41586-022-05172-4}.

\bibitem[Giannou et~al.(2023)Giannou, Rajput, Sohn, Lee, Lee, and Papailiopoulos]{looped}
Angeliki Giannou, Shashank Rajput, Jy-yong Sohn, Kangwook Lee, Jason~D. Lee, and Dimitris Papailiopoulos.
\newblock Looped transformers as programmable computers, 2023.
\newblock URL \url{https://arxiv.org/abs/2301.13196}.

\bibitem[Graves et~al.(2014)Graves, Wayne, and Danihelka]{neural-turing}
Alex Graves, Greg Wayne, and Ivo Danihelka.
\newblock Neural turing machines, 2014.
\newblock URL \url{https://arxiv.org/abs/1410.5401}.

\bibitem[Graves et~al.(2016)Graves, Wayne, Reynolds, Harley, Danihelka, Grabska-Barwińska, Colmenarejo, Grefenstette, Ramalho, Agapiou, Badia, Hermann, Zwols, Ostrovski, Cain, King, Summerfield, Blunsom, Kavukcuoglu, and Hassabis]{DNC}
Alex Graves, Greg Wayne, Malcolm Reynolds, Tim Harley, Ivo Danihelka, Agnieszka Grabska-Barwińska, Sergio~Gómez Colmenarejo, Edward Grefenstette, Tiago Ramalho, John Agapiou, Adrià~Puigdomènech Badia, Karl~Moritz Hermann, Yori Zwols, Georg Ostrovski, Adam Cain, Helen King, Christopher Summerfield, Phil Blunsom, Koray Kavukcuoglu, and Demis Hassabis.
\newblock {Hybrid computing using a neural network with dynamic external memory}.
\newblock \emph{Nature}, 538\penalty0 (7626):\penalty0 471--476, 2016.
\newblock ISSN 0028-0836.
\newblock \doi{10.1038/nature20101}.

\bibitem[Gupta \& Kembhavi(2022)Gupta and Kembhavi]{visual_program}
Tanmay Gupta and Aniruddha Kembhavi.
\newblock Visual programming: Compositional visual reasoning without training, 2022.

\bibitem[Hajali \& Budvytis(2023)Hajali and Budvytis]{fcps}
Patrick Hajali and Ignas Budvytis.
\newblock Function-constrained program synthesis.
\newblock In \emph{Advances in Neural Information Processing Systems, Workshop on robustness of zero/few-shot learning in foundation models}, 2023.

\bibitem[Hochreiter \& Schmidhuber(1997)Hochreiter and Schmidhuber]{LSTM}
Sepp Hochreiter and J\"{u}rgen Schmidhuber.
\newblock Long short-term memory.
\newblock \emph{Neural Comput.}, 9\penalty0 (8):\penalty0 1735–1780, nov 1997.
\newblock ISSN 0899-7667.
\newblock \doi{10.1162/neco.1997.9.8.1735}.
\newblock URL \url{https://doi.org/10.1162/neco.1997.9.8.1735}.

\bibitem[Kirillov et~al.(2023)Kirillov, Mintun, Ravi, Mao, Rolland, Gustafson, Xiao, Whitehead, Berg, Lo, Dollár, and Girshick]{segmentAnything}
Alexander Kirillov, Eric Mintun, Nikhila Ravi, Hanzi Mao, Chloe Rolland, Laura Gustafson, Tete Xiao, Spencer Whitehead, Alexander~C. Berg, Wan-Yen Lo, Piotr Dollár, and Ross Girshick.
\newblock Segment anything, 2023.

\bibitem[Li et~al.(2022)Li, Choi, Chung, Kushman, Schrittwieser, Leblond, Eccles, Keeling, Gimeno, Lago, Hubert, Choy, de~Masson~d'Autume, Babuschkin, Chen, Huang, Welbl, Gowal, Cherepanov, Molloy, Mankowitz, Robson, Kohli, de~Freitas, Kavukcuoglu, and Vinyals]{alphacode}
Yujia Li, David Choi, Junyoung Chung, Nate Kushman, Julian Schrittwieser, R{\'{e} }mi Leblond, Tom Eccles, James Keeling, Felix Gimeno, Agustin~Dal Lago, Thomas Hubert, Peter Choy, Cyprien de~Masson~d'Autume, Igor Babuschkin, Xinyun Chen, Po-Sen Huang, Johannes Welbl, Sven Gowal, Alexey Cherepanov, James Molloy, Daniel~J. Mankowitz, Esme~Sutherland Robson, Pushmeet Kohli, Nando de~Freitas, Koray Kavukcuoglu, and Oriol Vinyals.
\newblock Competition-level code generation with {AlphaCode}.
\newblock \emph{Science}, 378\penalty0 (6624):\penalty0 1092--1097, dec 2022.
\newblock \doi{10.1126/science.abq1158}.
\newblock URL \url{https://doi.org/10.1126%2Fscience.abq1158}.

\bibitem[Lindner et~al.(2023)Lindner, Kramár, Rahtz, McGrath, and Mikulik]{tracr}
David Lindner, János Kramár, Matthew Rahtz, Thomas McGrath, and Vladimir Mikulik.
\newblock Tracr: Compiled transformers as a laboratory for interpretability, 2023.
\newblock URL \url{https://arxiv.org/abs/2301.05062}.

\bibitem[Mankowitz et~al.(2023)Mankowitz, Michi, Zhernov, Gelmi, Selvi, Paduraru, Leurent, Iqbal, Lespiau, Ahern, Köppe, Millikin, Gaffney, Elster, Broshear, Gamble, Milan, Tung, Hwang, Cemgil, Barekatain, Li, Mandhane, Hubert, Schrittwieser, Hassabis, Kohli, Riedmiller, Vinyals, and Silver]{alphadev}
Daniel~J. Mankowitz, Andrea Michi, Anton Zhernov, Marco Gelmi, Marco Selvi, Cosmin Paduraru, Edouard Leurent, Shariq Iqbal, Jean-Baptiste Lespiau, Alex Ahern, Thomas Köppe, Kevin Millikin, Stephen Gaffney, Sophie Elster, Jackson Broshear, Chris Gamble, Kieran Milan, Robert Tung, Minjae Hwang, Taylan Cemgil, Mohammadamin Barekatain, Yujia Li, Amol Mandhane, Thomas Hubert, Julian Schrittwieser, Demis Hassabis, Pushmeet Kohli, Martin Riedmiller, Oriol Vinyals, and David Silver.
\newblock Faster sorting algorithms discovered using deep reinforcement learning.
\newblock \emph{Nature}, 618\penalty0 (7964):\penalty0 257--263, June 2023.
\newblock ISSN 1476-4687.
\newblock \doi{10.1038/s41586-023-06004-9}.
\newblock URL \url{https://doi.org/10.1038/s41586-023-06004-9}.

\bibitem[Mavaddat \& Parhami(1988)Mavaddat and Parhami]{SUBLEQ}
Farhad Mavaddat and Behrooz Parhami.
\newblock Urisc: The ultimate reduced instruction set computer.
\newblock \emph{International Journal of Electrical Engineering \& Education}, 25\penalty0 (4):\penalty0 327--334, 1988.
\newblock \doi{10.1177/002072098802500408}.
\newblock URL \url{https://doi.org/10.1177/002072098802500408}.

\bibitem[Nanda et~al.(2023)Nanda, Chan, Lieberum, Smith, and Steinhardt]{nanda2023progress}
Neel Nanda, Lawrence Chan, Tom Lieberum, Jess Smith, and Jacob Steinhardt.
\newblock Progress measures for grokking via mechanistic interpretability, 2023.

\bibitem[Olsson et~al.(2022)Olsson, Elhage, Nanda, Joseph, DasSarma, Henighan, Mann, Askell, Bai, Chen, Conerly, Drain, Ganguli, Hatfield-Dodds, Hernandez, Johnston, Jones, Kernion, Lovitt, Ndousse, Amodei, Brown, Clark, Kaplan, McCandlish, and Olah]{olsson2022context}
Catherine Olsson, Nelson Elhage, Neel Nanda, Nicholas Joseph, Nova DasSarma, Tom Henighan, Ben Mann, Amanda Askell, Yuntao Bai, Anna Chen, Tom Conerly, Dawn Drain, Deep Ganguli, Zac Hatfield-Dodds, Danny Hernandez, Scott Johnston, Andy Jones, Jackson Kernion, Liane Lovitt, Kamal Ndousse, Dario Amodei, Tom Brown, Jack Clark, Jared Kaplan, Sam McCandlish, and Chris Olah.
\newblock In-context learning and induction heads.
\newblock \emph{Transformer Circuits Thread}, 2022.
\newblock https://transformer-circuits.pub/2022/in-context-learning-and-induction-heads/index.html.

\bibitem[Ouyang et~al.(2022)Ouyang, Wu, Jiang, Almeida, Wainwright, Mishkin, Zhang, Agarwal, Slama, Ray, Schulman, Hilton, Kelton, Miller, Simens, Askell, Welinder, Christiano, Leike, and Lowe]{RLHF}
Long Ouyang, Jeff Wu, Xu~Jiang, Diogo Almeida, Carroll~L. Wainwright, Pamela Mishkin, Chong Zhang, Sandhini Agarwal, Katarina Slama, Alex Ray, John Schulman, Jacob Hilton, Fraser Kelton, Luke Miller, Maddie Simens, Amanda Askell, Peter Welinder, Paul Christiano, Jan Leike, and Ryan Lowe.
\newblock Training language models to follow instructions with human feedback, 2022.
\newblock URL \url{https://arxiv.org/abs/2203.02155}.

\bibitem[Popescu et~al.(2009)Popescu, Balas, Perescu-Popescu, and Mastorakis]{mlp}
Marius-Constantin Popescu, Valentina Balas, Liliana Perescu-Popescu, and Nikos Mastorakis.
\newblock Multilayer perceptron and neural networks.
\newblock \emph{WSEAS Transactions on Circuits and Systems}, 8, 07 2009.

\bibitem[Romera-Paredes et~al.(2024)Romera-Paredes, Barekatain, Novikov, Balog, Kumar, Dupont, Ruiz, Ellenberg, Wang, Fawzi, Kohli, and Fawzi]{FunSearch}
Bernardino Romera-Paredes, Mohammadamin Barekatain, Alexander Novikov, Matej Balog, M.~Pawan Kumar, Emilien Dupont, Francisco J.~R. Ruiz, Jordan~S. Ellenberg, Pengming Wang, Omar Fawzi, Pushmeet Kohli, and Alhussein Fawzi.
\newblock Mathematical discoveries from program search with large language models.
\newblock \emph{Nature}, 625\penalty0 (7995):\penalty0 468--475, January 2024.
\newblock ISSN 1476-4687.
\newblock \doi{10.1038/s41586-023-06924-6}.
\newblock URL \url{https://doi.org/10.1038/s41586-023-06924-6}.

\bibitem[Schick et~al.(2023)Schick, Dwivedi-Yu, Dessì, Raileanu, Lomeli, Zettlemoyer, Cancedda, and Scialom]{toolformer}
Timo Schick, Jane Dwivedi-Yu, Roberto Dessì, Roberta Raileanu, Maria Lomeli, Luke Zettlemoyer, Nicola Cancedda, and Thomas Scialom.
\newblock Toolformer: Language models can teach themselves to use tools, 2023.

\bibitem[Silver et~al.(2018)Silver, Hubert, Schrittwieser, Antonoglou, Lai, Guez, Lanctot, Sifre, Kumaran, Graepel, Lillicrap, Simonyan, and Hassabis]{alphazero}
David Silver, Thomas Hubert, Julian Schrittwieser, Ioannis Antonoglou, Matthew Lai, Arthur Guez, Marc Lanctot, Laurent Sifre, Dharshan Kumaran, Thore Graepel, Timothy Lillicrap, Karen Simonyan, and Demis Hassabis.
\newblock A general reinforcement learning algorithm that masters chess, shogi, and go through self-play.
\newblock \emph{Science}, 362\penalty0 (6419):\penalty0 1140--1144, 2018.
\newblock \doi{10.1126/science.aar6404}.
\newblock URL \url{https://www.science.org/doi/abs/10.1126/science.aar6404}.

\bibitem[Touvron et~al.(2023)Touvron, Lavril, Izacard, Martinet, Lachaux, Lacroix, Rozière, Goyal, Hambro, Azhar, Rodriguez, Joulin, Grave, and Lample]{touvron2023llama}
Hugo Touvron, Thibaut Lavril, Gautier Izacard, Xavier Martinet, Marie-Anne Lachaux, Timothée Lacroix, Baptiste Rozière, Naman Goyal, Eric Hambro, Faisal Azhar, Aurelien Rodriguez, Armand Joulin, Edouard Grave, and Guillaume Lample.
\newblock Llama: Open and efficient foundation language models, 2023.

\bibitem[Vaswani et~al.(2017)Vaswani, Shazeer, Parmar, Uszkoreit, Jones, Gomez, Kaiser, and Polosukhin]{transformer}
Ashish Vaswani, Noam Shazeer, Niki Parmar, Jakob Uszkoreit, Llion Jones, Aidan~N. Gomez, Lukasz Kaiser, and Illia Polosukhin.
\newblock Attention is all you need, 2017.
\newblock URL \url{https://arxiv.org/abs/1706.03762}.

\bibitem[Verenich et~al.(2020)Verenich, Velasquez, Murshed, and Hussain]{feature-reuse}
Edward Verenich, Alvaro Velasquez, M.~G.~Sarwar Murshed, and Faraz Hussain.
\newblock The utility of feature reuse: Transfer learning in data-starved regimes, 2020.

\bibitem[Wang et~al.(2023)Wang, Xie, Jiang, Mandlekar, Xiao, Zhu, Fan, and Anandkumar]{voyager}
Guanzhi Wang, Yuqi Xie, Yunfan Jiang, Ajay Mandlekar, Chaowei Xiao, Yuke Zhu, Linxi Fan, and Anima Anandkumar.
\newblock Voyager: An open-ended embodied agent with large language models, 2023.

\bibitem[Weiss et~al.(2021)Weiss, Goldberg, and Yahav]{RASP}
Gail Weiss, Yoav Goldberg, and Eran Yahav.
\newblock Thinking like transformers, 2021.
\newblock URL \url{https://arxiv.org/abs/2106.06981}.

\bibitem[Zaremba \& Sutskever(2014)Zaremba and Sutskever]{learning-to-execute}
Wojciech Zaremba and Ilya Sutskever.
\newblock Learning to execute, 2014.
\newblock URL \url{https://arxiv.org/abs/1410.4615}.

\bibitem[Zhou et~al.(2022)Zhou, Nova, Larochelle, Courville, Neyshabur, and Sedghi]{algo-in-context}
Hattie Zhou, Azade Nova, Hugo Larochelle, Aaron Courville, Behnam Neyshabur, and Hanie Sedghi.
\newblock Teaching algorithmic reasoning via in-context learning, 2022.
\newblock URL \url{https://arxiv.org/abs/2211.09066}.

\bibitem[Zhou et~al.(2023)Zhou, Bradley, Littwin, Razin, Saremi, Susskind, Bengio, and Nakkiran]{length-generalization}
Hattie Zhou, Arwen Bradley, Etai Littwin, Noam Razin, Omid Saremi, Josh Susskind, Samy Bengio, and Preetum Nakkiran.
\newblock What algorithms can transformers learn? a study in length generalization, 2023.

\end{thebibliography}
\bibliographystyle{iclr2024_conference}

\end{document}